\documentclass{article}

    \usepackage[nonatbib]{neurips_2021}

\usepackage[utf8]{inputenc} % allow utf-8 input
\usepackage[T1]{fontenc}    % use 8-bit T1 fonts
\usepackage{hyperref}       % hyperlinks
\usepackage{url}            % simple URL typesetting
\usepackage{booktabs}       % professional-quality tables
\usepackage{amsfonts}       % blackboard math symbols
\usepackage{nicefrac}       % compact symbols for 1/2, etc.
\usepackage{microtype}      % microtypography
\usepackage{xcolor}         % colors
\usepackage{biblatex} %Imports biblatex package

\usepackage[colorinlistoftodos]{todonotes}
\usepackage{graphicx}
\usepackage{subfig}

\title{Transformer based trajectory prediction}

\author{Aleksey Postnikov$^{1,2}$ \and Aleksander Gamayunov$^{1,2}$ \and Gonzalo Ferrer$^{2}$%
        \thanks{\textsuperscript{1} The authors are with the Sberbank Robotics Laboratory, Moscow, Russia.
                    {\tt\small \{postnikov.a.l,gamayunov.a.r\}@sberbank.ru}.
               }% <-this % stops a space
        \thanks{\textsuperscript{2}Skolkovo Institute of Science and Technology, Moscow, Russia.
                    {\tt\small g.ferrer@skoltech.ru}.
           }    
    }
\begin{document}

\maketitle

\begin{abstract}
To plan a safe and efficient route, an autonomous vehicle should anticipate future motions of other agents around it. Motion prediction is an extremely challenging task which recently gained significant attention of the research community. 

In this work, we present a simple and yet strong baseline for uncertainty aware motion prediction based purely on transformer neural networks, which has shown its effectiveness in conditions of domain change. While being easy-to-implement, the proposed approach achieves competitive performance and ranks 1$^{st}$ on the 2021 Shifts Vehicle Motion Prediction Competition.
\end{abstract}
\section{Introduction}
 
% \todo{ unfinished }

% 1) What is the problem
% (by an actor we mean any vehicle, pedestrian, bicycle or other potentially dynamic object)
Driving a car is a complex activity that requires drivers to understand the involved multi-actor scenes in real time and actions in a rapidly changing environment in a fraction of second. To be able to fully rely on autonomous vehicles to drive autonomously, desirable to correctly assess the confidence of the algorithm in its predictions, including situations under the condition of distributional shifts, e.g. in a unseen (new to algorithms) roads, cities, countries.

% 2) Why is it important

In order to fully rely at autonomous vehicles, it is necessary to be confident of a high level of generalization of all algorithms used for autonomous driving.

The   motivation to understand and predict human motion  is  immense  and  it  has  a  deep  impact  in related  topics,  such  as,  decision  making,  path  planning, autonomous navigation, surveillance, tracking, scene under-standing, anomaly detection, etc.

% 3) Why is it hard

The  problem  of  forecasting  where  cars  will  be  in the near future is, however, ill-posed by nature: human beings tend  to  be  unpredictable  on  their  decisions  and  car driving is neither exempt of it. These random nature in motion brings an open challenge to prediction algorithms, where algorithms are desired to be accurate and correctly grasp the uncertainty associated with their predictions.

% 4) what do we do?

The contributions of this work are summarized as follows: 1) we propose a unified transformer-based motion prediction framework for both multi-modal trajectory prediction and uncertainty estimation. 2) Our proposed approach achieves state-of-the-art performance, and ranks 1$^{st}$ on the Shifts Vehicle Motion Prediction Competition.

% Driving a car is a complex activity that requires drivers to understand the involved multi-actor(by an actor we mean any vehicle, pedestrian, bicycle or other potentially dynamic object) scenes in real time and actions in a rapidly changing environment in a split second. 
% Human behavior, including the behavior of a person driving a car, is often extremely multimodal and difficult to predict. Therefore, many modern approaches [], in addition to predicting the trajectories directly, predict uncertainty.

% Unfortunately, humans are infamously ill-fitted for the task, as sadly corroborated by grim road statistics that often worsen year after year. Traffic accidents were the number four cause of death in the US in 2015, accounting for more than 5\% of the total [31]. In addition, despite large investments by governments and progress made in traffic safety technologies, in the US the year 2017 was still one of the deadliest years for motorists in the past decade [33]. Moreover, human error is responsible for up to 94\% of crashes [41], suggesting that removing the unreliable human factor could potentially save hundreds of thousands of lives and tens of billions of dollars in accident-related damages and medical expenses [6].

\section{Related Work}

% \todo{ start and finish} 

% motion prediction - cnn - CV - vit. 

The motion prediction task is one of the most important in the field of autonomous driving and has recently attracted a lot of attention from both academia and industry \cite{alahi2016social} \cite{lee2017desire} \cite{gupta2018social} \cite{chai2019multipath} \cite{salzmann2020trajectron++} \cite{postnikov2020hsfm}.

Broadly, modern motion prediction methods can be divided in two classes: 

1) Models where scene context information are processed from vectorized maps (HD maps) \cite{gao2020vectornet} \cite{liang2020learning}.

2) Models where high-definition maps and surroundings of each vehicle in cars’s vicinity rasterized to image representation, thus providing complete context and information necessary for
accurate prediction of future trajectory \cite{djuric2020uncertainty} \cite{postnikov2021covariancenet} \cite{fang2020tpnet}.

Recently models based at transformers architectures , have shown theirs applicability both at computer vision tasks \cite{dosovitskiy2020image} \cite{liu2021swin}, and at sequence to sequence tasks \cite{radford2018improving}, \cite{devlin2018bert}, which opens a high potential of applicability Transformer based approaches for motion prediction task.
\section{Method}

We assume that object detections and tracks are provided by perception stack (running on the Yandex self driving car (SDC) fleet \cite{malinin2021shifts}) and focus only on the motion prediction.

%  The goal of the task is to predict the movement trajectory of vehicles at time
The proposed method goal is to predict the most-likely movement trajectory of vehicles at time $T \in (0, 5]$  seconds in the future and model's scalar uncertainty estimates, which can later be used in subsequent SDC pipeline algorithms as an estimate of the forecast uncertainty with a scene context that is particularly familiar or low risk in the case of low estimated uncertainty, or unfamiliar or high risk in the case of high estimated uncertainty.
% Task of the model is to predict most-likely trajectories of an agent for the next T=5 seconds in the future and model's scalar uncertainty

In this section we describe the architecture of our model, the loss function used for training and implementation details.

\begin{figure}[t!]
    \includegraphics[width=14cm]{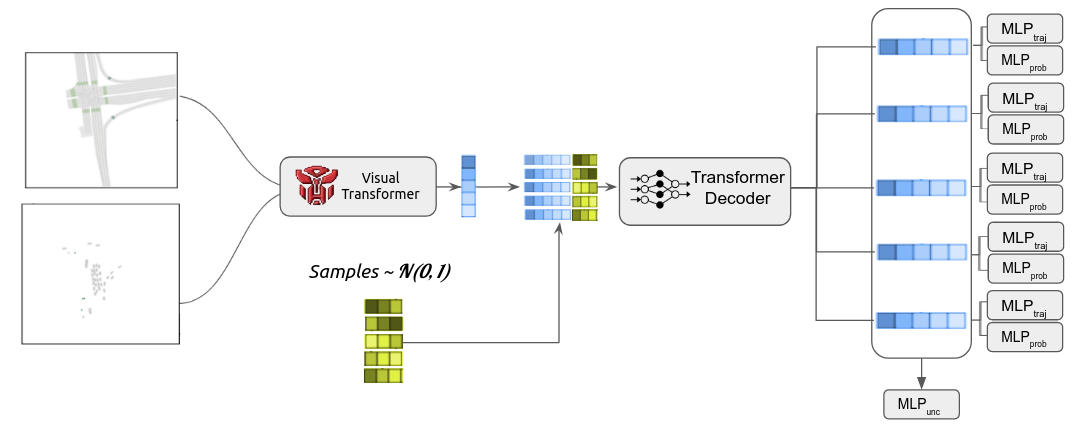}
    \caption{Transformer based motion prediction architecture. Multi-dimensional rasterized scene encoded by visual transformer, latent scene representation repeated $K$ times, according to number of trajectories to be predicted, concatenated to $S \sim \mathcal{N}(0,1)$. Latent states transformed to predicted trajectories, their probabilities and general model uncertainty with transforer-based decoder and Multi Layer Perceptrons.}
    \label{fig::crop_rot}
\end{figure}

% successful system also
% needs to account for their inherent multimodal nature
The future is ambiguous and human motion is unpredictable and multimodal by nature. In order to account for such multimodal nature, we aim to produce up to K=5 different hypotheses (proposals) and their probabilities for the future trajectory which will be evaluated against the ground truth trajectory.

\subsection{Input representation}
Context information about the state of dynamic objects (i.e., vehicles, pedestrians), described by its position, velocity, linear acceleration, and orientation together with context information about the HD map including lane information (e.g., traffic direction, lane priority, speed limit, traffic light association), road boundaries, crosswalks, and traffic light states are rasterized into multi-channel images which are passed to transformer based model.

\subsection{Model architecture}
Our method can be represented as an image-based regression, model architecture shown at Fig \ref{fig::crop_rot} consists of two main stages :
\begin{itemize}

%  Past observation, consisted of positions, velocities, and accelerations of all dynamical objects detected around and rasterized HD map of environment consisted of road polygons, lanes, crossroads information grouped together as a 17 channel image and passed to image encoder.
\item
Transformer-based image processing encoder, namely modified ViT\cite{dosovitskiy2020image}  (modified in accordance to process multi-channel images) acts as a current state estimation for single vehicle. ViT uses multi-head self-attention\cite{vaswani2017attention} removing image-specific inductive biases compared to CNN approaches and self-attention layers in ViT allows it to integrate information globally across the entire image so ViT doesn't suffer from lacks of a global understanding of the images.

\item Transformer based decoder witch predicts $K$ different hypothesis (proposals) for the future trajectory with the corresponding confidence values $c^{1..5}$ which are normalized using softmax operator such that $\sum c^{1..5} = 1 $ . Apart of predicting multi-trajectory plans proposed model predicts overall scene uncertainty score, which is described in more details later.
\end{itemize}

Multi channel images are initially split into fixed-size patches and processed further by to visual transformer model. Dense encoded latent state produced by ViT later repeated N times, according to number of desired trajectories to be predicted.

Each of $K$ latent state concatenated with $S \sim \mathcal{N}(0,1)$, samples from Normal distribution, which is according to our internal experiments gives minor improvements in metrics comparing to sinusoidal Positional Encoding \cite{vaswani2017attention}, that can be interpreted by the absence of a relative or absolute positional correlation between sequential states. 
On the other hand samples from Normal distribution, concatenated with repeated latent state, helps decoder to transform repeated latent state to more diverse trajectories.

%  Transformers can attend to complete sequences  
At the same time, property of multi-head attention attend globally, therefore, learning long-range relationships provides more opportunities for a correct assessment of uncertainties.
% \todo{ talk about prediction uncertainties and why transfomers is good at that}

\subsection{Loss function}

\begin{figure}[t]
    \centering
    \subfloat[\centering Examples of multi-modality of predicted trajectories ]{{\includegraphics[width=6.5cm]{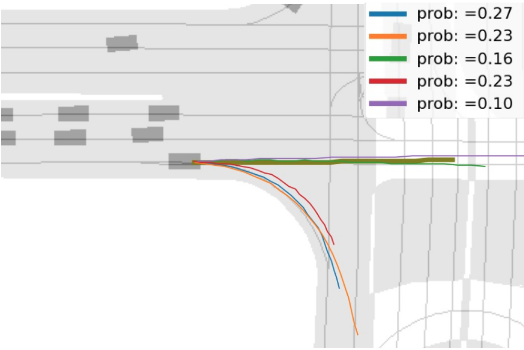} }}%
    \qquad
    \subfloat[\centering Negative example with high average displacement error.] {{\includegraphics[width=6.5cm]{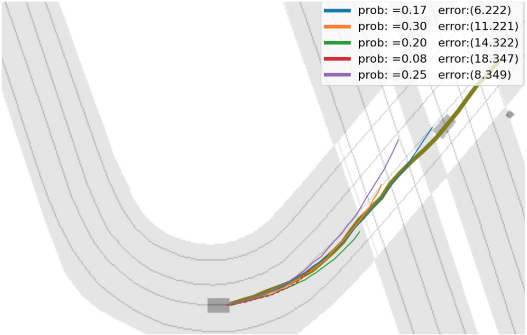} }}%
    \caption{Qualitative results of Transformer based trajectory prediction on  Shifts Vehicle Motion Prediction validation set. Bold olive line represents ground truth trajectory, other colored lines represents trajectories predicted by proposed model, legend shows probability of each predicted trajectory, and ADE (error) of that trajectory}%
    \label{fig:example}%
\end{figure}

% One of the most popular solution would be to use a Mean Squared Error (MSE) loss. However, this loss does not allow a probabilistic modelling of multi-modal hypotheses and it showed poor performance in our preliminary experiments. 
% Instead, 
We model possible future trajectories as the mixture of $K$ Gaussian distributions, as it is allows model to predict multi-modal distribution,  comparing to widely spread ADE loss, examples of model's predictions are shown at Fig \ref{fig:example}. 
In this case our network outputs the means positions of the Gaussians $\hat{x}_i^j$ while we fix the covariance of every Gaussian in the mixture to be equal to the identity matrix $\sigma = I$, and for each trajectory model predicts trajectory probability $c^k$ which are normalized using softmax operator such that $\sum c^k = 1$.
Then, given predicted $\hat{x}_i^j, \sigma_i^k, c_i^k$ for the loss function we can use negative log-likelihood (NLL) of mixture of Gaussians defined by the predicted proposals given the ground truth coordinates $X^{gt}$. 

\begin{equation}
    X^{gt} = [(x_1,y_1), ... ,(x_T,y_T)]
\end{equation}
\begin{equation}
    \hat{X} = [(\hat{x}^k_1,\hat{y}^k_1), ... ,(\hat{x}^k_T,\hat{y}^k_T)] , k = 1, ..., K
\end{equation}
where T is a prediction horizon, K - number of hypotheses

We compute negative log probability of the ground truth trajectory under the predicted mixture of Gaussians with the means equal to the predicted trajectories and the identity matrix I as covariance:

\begin{equation}
    L_{pose} = -log \sum^K_{k=1} c_k \mathcal{N} (X^{gt}; \mu = \hat{X}; \sigma=I )
\end{equation}

In order to evaluate model uncertainty, we propose to use second loss which is basically Root Mean Squared Error between predicted uncertainty measure and trajectory NLL value.

\begin{equation}
    L_{uncertanty} = RMSE(L_{pose}, \hat{U})
\end{equation}

where U - predicted uncertainty score,  RMSE - Root Mean Squared Error function. 

\subsection{Implementation details}

The output of our model is $K = 5$ trajectories, each  containing $T=25$ two dimensional coordinates. 
We train our model using AdamW \cite{loshchilov2017decoupled} optimizing for 40 full epochs of training set provided by shifts vehicle motion prediction challenge \cite{malinin2021shifts}, and SGD for yet other 40 full epochs.  
We use a learning rate of $10^{-4}$ for AdamW \cite{loshchilov2017decoupled} optimizer, weight decay $10^{-2}$ and a batch size of 1024. 
We use cosine annealing scheduler for AdamW\cite{loshchilov2017decoupled}  with warm-up. 
For SGD optimizer we use initial learning rate of $10^{-3}$ and cosine annealing scheduler with warm-up and restarts.

As an encoder we utilise ViT-Base model, with layers number $L=12$  and Hidden size $D=768$, we modify final layer of ViT to output 512 features. dimension of samples from normal distribution is 8, Transformer decoder consist of $L=8$ layers with hidden size $D=2048$, number of heads $N=8$.

% \todo {explain shapes}

We training our model with ViT \cite{dosovitskiy2020image} backbone for 7 days on a two NVIDIA V100 GPU with 2*32Gb VRAM in total.

\section{Experiments}

We evaluate the effectiveness of Transformer based trajectory prediction method on the Shifts Vehicle Motion Prediction Challenge \cite{malinin2021shifts}. 
As shown in Table 1, our method ranks 1$^{st}$ on the leaderboard. 
The Shifts Motion Prediction Challenge main metric is R-AUC cNLL\cite{malinin2021shifts}, which provides a full picture of the models performance incorporating both uncertainty estimation and predicted trajectories accuracy. It can be seen that despite the fact that accoring to R-AUC cNLL metric second method gives slightly better trajectories accuracy in terms of ADE, FDE, cNLL, our transformer-based approach produces more reliable uncertainties which reflect in better overall R-AUC cNLL.

\begin{table}[h]
    \centering
    \caption{Top 4 entries of the Shifts Vehicle Motion Prediction Challenge.}
    \begin{tabular}{||c|c|c|c|c||}
        \hline
        Team Name & \textbf{R-AUC cNLL}   &  cNLL & Min ADE (k=5) & Min FDE (k=5) \\
        \hline\hline
        HOME   & 3.45 &  24.63 &  0.54 & 1.07	      \\
        \hline
        Alexey \& Dmitry & 2.62 & 15.59 & 0.50 & 0.94 \\
        \hline 
        NTU CMLab Mira & 8.63 & 61.86 & 0.80 & 1.72   \\
        \hline \hline
        SBteam (ours) & \textbf{2.57} & 15.67  &  0.53 & 1.01 \\
        \hline
    \end{tabular}
    
    \label{tab:my_label}
\end{table}
\section{Conclusions}
In this work, we propose an fully tranformer-based trajectory prediction model. It outperforms previous cnn-based and graph-based methods at the task of uncertainty aware motion prediction. Our model achieves state-of-the-art performance and ranks 1$^{st}$ on the Shifts Motion Prediction Challenge.

\printbibliography

@inproceedings{vaswani2017attention,
  title={Attention is all you need},
  author={Vaswani, Ashish and Shazeer, Noam and Parmar, Niki and Uszkoreit, Jakob and Jones, Llion and Gomez, Aidan N and Kaiser, {\L}ukasz and Polosukhin, Illia},
  booktitle={Advances in neural information processing systems},
  pages={5998--6008},
  year={2017}
}

@article{dosovitskiy2020image,
  title={An image is worth 16x16 words: Transformers for image recognition at scale},
  author={Dosovitskiy, Alexey and Beyer, Lucas and Kolesnikov, Alexander and Weissenborn, Dirk and Zhai, Xiaohua and Unterthiner, Thomas and Dehghani, Mostafa and Minderer, Matthias and Heigold, Georg and Gelly, Sylvain and others},
  journal={arXiv preprint arXiv:2010.11929},
  year={2020}
}

@article{loshchilov2017decoupled,
  title={Decoupled weight decay regularization},
  author={Loshchilov, Ilya and Hutter, Frank},
  journal={arXiv preprint arXiv:1711.05101},
  year={2017}
}

@article{malinin2021shifts,
  title={Shifts: A dataset of real distributional shift across multiple large-scale tasks},
  author={Malinin, Andrey and Band, Neil and Chesnokov, German and Gal, Yarin and Gales, Mark JF and Noskov, Alexey and Ploskonosov, Andrey and Prokhorenkova, Liudmila and Provilkov, Ivan and Raina, Vatsal and others},
  journal={arXiv preprint arXiv:2107.07455},
  year={2021}
}

@inproceedings{alahi2016social,
  title={Social lstm: Human trajectory prediction in crowded spaces},
  author={Alahi, Alexandre and Goel, Kratarth and Ramanathan, Vignesh and Robicquet, Alexandre and Fei-Fei, Li and Savarese, Silvio},
  booktitle={Proceedings of the IEEE conference on computer vision and pattern recognition},
  pages={961--971},
  year={2016}
}

@inproceedings{lee2017desire,
  title={Desire: Distant future prediction in dynamic scenes with interacting agents},
  author={Lee, Namhoon and Choi, Wongun and Vernaza, Paul and Choy, Christopher B and Torr, Philip HS and Chandraker, Manmohan},
  booktitle={Proceedings of the IEEE Conference on Computer Vision and Pattern Recognition},
  pages={336--345},
  year={2017}
}

@inproceedings{gupta2018social,
  title={Social gan: Socially acceptable trajectories with generative adversarial networks},
  author={Gupta, Agrim and Johnson, Justin and Fei-Fei, Li and Savarese, Silvio and Alahi, Alexandre},
  booktitle={Proceedings of the IEEE Conference on Computer Vision and Pattern Recognition},
  pages={2255--2264},
  year={2018}
}

@article{chai2019multipath,
  title={Multipath: Multiple probabilistic anchor trajectory hypotheses for behavior prediction},
  author={Chai, Yuning and Sapp, Benjamin and Bansal, Mayank and Anguelov, Dragomir},
  journal={arXiv preprint arXiv:1910.05449},
  year={2019}
}

@inproceedings{salzmann2020trajectron++,
  title={Trajectron++: Dynamically-feasible trajectory forecasting with heterogeneous data},
  author={Salzmann, Tim and Ivanovic, Boris and Chakravarty, Punarjay and Pavone, Marco},
  booktitle={Computer Vision--ECCV 2020: 16th European Conference, Glasgow, UK, August 23--28, 2020, Proceedings, Part XVIII 16},
  pages={683--700},
  year={2020},
  organization={Springer}
}

@inproceedings{gao2020vectornet,
  title={Vectornet: Encoding hd maps and agent dynamics from vectorized representation},
  author={Gao, Jiyang and Sun, Chen and Zhao, Hang and Shen, Yi and Anguelov, Dragomir and Li, Congcong and Schmid, Cordelia},
  booktitle={Proceedings of the IEEE/CVF Conference on Computer Vision and Pattern Recognition},
  pages={11525--11533},
  year={2020}
}

@inproceedings{liang2020learning,
  title={Learning lane graph representations for motion forecasting},
  author={Liang, Ming and Yang, Bin and Hu, Rui and Chen, Yun and Liao, Renjie and Feng, Song and Urtasun, Raquel},
  booktitle={European Conference on Computer Vision},
  pages={541--556},
  year={2020},
  organization={Springer}
}

@article{postnikov2021covariancenet,
  title={CovarianceNet: Conditional generative model for correct covariance prediction in human motion prediction},
  author={Postnikov, Aleksey and Gamayunov, Aleksander and Ferrer, Gonzalo},
  journal={arXiv preprint arXiv:2109.02965},
  year={2021}
}

@inproceedings{djuric2020uncertainty,
  title={Uncertainty-aware short-term motion prediction of traffic actors for autonomous driving},
  author={Djuric, Nemanja and Radosavljevic, Vladan and Cui, Henggang and Nguyen, Thi and Chou, Fang-Chieh and Lin, Tsung-Han and Singh, Nitin and Schneider, Jeff},
  booktitle={Proceedings of the IEEE/CVF Winter Conference on Applications of Computer Vision},
  pages={2095--2104},
  year={2020}
}

@article{postnikov2020hsfm,
  title={HSFM-Sigmann: Combining a Feedforward Motion Prediction Network and Covariance Prediction},
  author={Postnikov, Aleksey and Gamayunov, Aleksander and Ferrer, Gonzalo},
  journal={arXiv preprint arXiv:2009.04299},
  year={2020}
}

@inproceedings{fang2020tpnet,
  title={Tpnet: Trajectory proposal network for motion prediction},
  author={Fang, Liangji and Jiang, Qinhong and Shi, Jianping and Zhou, Bolei},
  booktitle={Proceedings of the IEEE/CVF Conference on Computer Vision and Pattern Recognition},
  pages={6797--6806},
  year={2020}
}

@article{liu2021swin,
  title={Swin transformer: Hierarchical vision transformer using shifted windows},
  author={Liu, Ze and Lin, Yutong and Cao, Yue and Hu, Han and Wei, Yixuan and Zhang, Zheng and Lin, Stephen and Guo, Baining},
  journal={arXiv preprint arXiv:2103.14030},
  year={2021}
}

@article{devlin2018bert,
  title={Bert: Pre-training of deep bidirectional transformers for language understanding},
  author={Devlin, Jacob and Chang, Ming-Wei and Lee, Kenton and Toutanova, Kristina},
  journal={arXiv preprint arXiv:1810.04805},
  year={2018}
}

@article{radford2018improving,
  title={Improving language understanding by generative pre-training},
  author={Radford, Alec and Narasimhan, Karthik and Salimans, Tim and Sutskever, Ilya},
  year={2018}
}

\end{document}